\documentclass{ecai}
\usepackage{graphicx}
\usepackage{latexsym}
\usepackage[linesnumbered,ruled,vlined]{algorithm2e}

\begin{document}
\begin{frontmatter}

\title{Zoom and Shift is All You Need}

\author[A]{\fnms{}~\snm{Jiahao Qin}\thanks{qjh2020@liverpool.ac.uk}\orcid{https://orcid.org/0000-0002-0551-4647}}



\address[A]{Xi’an Jiaotong-Liverpool University}

\begin{abstract}
Feature alignment serves as the primary mechanism for fusing multimodal data. We put forth a feature alignment approach that achieves full integration of multimodal information. This is accomplished via an alternating process of shifting and expanding feature representations across modalities to obtain a consistent unified representation in a joint feature space. The proposed technique can reliably capture high-level interplay between features originating from distinct modalities. Consequently, substantial gains in multimodal learning performance are attained. Additionally, we demonstrate the superiority of our approach over other prevalent multimodal fusion schemes on a range of tasks. Extensive experimental evaluation conducted on multimodal datasets comprising time series, image, and text demonstrates that our method achieves state-of-the-art results.
\end{abstract}

\end{frontmatter}

\section{Introduction}
Multimodal data, such as images, audio, and text, is ubiquitous in the real world. The data for each model has different eigenvectors, However, each modality has its own set of eigenvectors, which are located in different subspaces \cite{lahat_multimodal_2015}. Consequently, they have different distributions and statistical properties, resulting in vectors with similar semantics being represented differently in different subspaces. Therefore, vectors with similar semantics represented in different subspaces will be completely different. This phenomenon is commonly known as the heterogeneity gap.

The heterogeneity gap is a significant issue as it can hinder the integrated use of multimodal data by subsequent machine learning modules \cite{lahat_multimodal_2015}. The traditional approach to addressing this issue is to transform data from different modalities into the same vector space, for example, through feature fusion or dimensionality reduction methods. 
 However, the disadvantage of this method is that the modal fusion is difficult, it is not easy to realize the cross-modal learning model, and the model cannot be extended to the single-modal scene \cite{cho_unifying_2021}.

 This method has been widely used in multimodal data processing and has achieved certain results. For example, in text-image retrieval, this method can be used to map text descriptions and image features into a common subspace to achieve cross-modal retrieval \cite{cho_unifying_2021}. And in audio-image retrieval, this method can also be used to map audio and image features into a common subspace to achieve cross-modal retrieval \cite{yang2022multimodal} \cite{yuan2022exploring} \cite{zhang2007cross}.

 However, there are some problems with this approach. For example, due to the heterogeneous gap, the mapping relationship between different modes is difficult to learn, which leads to the poor effect of the mapping function \cite{guo_deep_2019}. In addition, due to the high dimensionality of the feature space, the computational complexity of the learning mapping function is also high \cite{houle_can_2010}. Therefore, how to effectively learn the mapping function and reduce the computational complexity is a problem that this method needs to solve.

We design a new alignment method: Alternative Telescopic Displacement(ATD), to overcome the above problems. We scale, rotate, and displace different feature spaces when fusing information features from different modalities. Displacement mapping is uncomplicated, and it can reduce the complexity of the model and prevent the gradient from disappearing.  Therefore, ATD could not only fully integrate the feature information between modalities and within modalities but also could reduce overfitting and prevent gradient disappearance. 

In addition, if a larger feature space is directly used to represent the features of multiple modalities, higher feature dimensions and more complex calculations will be inevitable. Before displacement mapping, we alternately select the features of one of the modalities for scaling and rotation transformation, which can reduce the dimensionality of the final alignment space. Also, we alternate between operations of stretching and rotating to add essential information about other modality features. In this way, we greatly avoid information loss.

\section{Related works}

Multimodal learning leverages multiple data types like text, audio, and images to overcome limitations of single modality approaches. Unimodal methods yield insufficient representations as they cannot access information from other modalities and lack cross-modal interactions\cite{lappe_cortical_2008,liang_expanding_2022,chen_cross-modal_2022}. In contrast, multimodal integration provides more comprehensive inputs to improve model accuracy and robustness\cite{baltrusaitis_multimodal_2019}.

Multimodal techniques have advanced many fields where multiple modalities exist naturally. In computer vision, fusing x-ray, CT, and MRI images boosts medical classification\cite{subbiah_parvathy_optimal_2020}. For target recognition, integrating camera and LiDAR data enhances robustness over using RGB or 3D alone\cite{melotti_multimodal_2020}. In NLP, scaling up pre-trained dual-modality models with billions of image-text pairs improves vision and language tasks\cite{jia_scaling_2021}.

Furthermore, multimodal methods excel for inherently cross-modal tasks like audio-visual speech recognition. Models fusing mouth movements, waveforms, and lip images outperform unimodal approaches\cite{paraskevopoulos_multiresolution_2020}\cite{tatulli_feature_2017}. Video classification also benefits from jointly modeling frames, subtitles, and audio\cite{mroueh_deep_2015}.

A key advantage of multimodality is overcoming deficiencies of unimodal methods. Unimodal approaches utilize only one modality, lacking information from other modalities and their interactions\cite{lappe_cortical_2008,liang_expanding_2022,chen_cross-modal_2022}. This can yield insufficient representations. Multimodal integration provides more comprehensive inputs by consolidating information across modalities, enhancing model performance.

Multimodal learning is widely used for vision, speech, and text tasks\cite{baltrusaitis_multimodal_2019} . For example, combining vision and audio data improves deep network performance for depression classification\cite{meng_depression_2013}. In sentiment analysis, fusing text, images, and audio extracts more informative features than using just text\cite{soleymani_survey_2017}.

In short, multimodal learning better leverages real-world data with natural synergies between modalities. Consolidating text, visual, and audio data addresses limitations of single data types. This flexibility enables multimodal techniques to advance a diverse range of tasks across disciplines.

\section{Multimodal model with Alternating Telescopic Displacement modules}

\subsection{Overview}

This work proposes a multimodal fusion framework to enable flexible bimodal learning across modalities with both shared and unique characteristics. The framework comprises three key modules designed to model, calibrate, and fuse multimodal features in order to address performance issues caused by heterogeneity between modalities:

Alternative Telescopic Displacement (ATD) Encoders: The framework utilizes separate ATD encoders based entirely on transformer-like structures to extract informative features from each distinct modality, including time series, images, and text. The uniform transformer architecture facilitates capturing nuanced modality-specific representations despite differences between data types.

ATD Guide Module: While the ATD encoders transform each modality into a standardized feature space, the distributions of the output values differ across modalities due to heterogeneity. To address this, the ATD Guide module learns scaling factors conditioned on the modality type to calibrate the output distributions. This bridges representation gaps between the modalities to enable unified multimodal fusion.

ATD Fusion Module: This module aligns and fuses the calibrated encoded features from the different modalities through an attention mechanism. The aligned multimodal representations can then be fed into downstream prediction tasks.

Together, the proposed model is a multi-module transformer-based architecture provides a flexible framework for bimodal learning and fusion while accounting for cross-modal heterogeneity. The ATD encoders leverage the uniform structure to model uniqueness of each modality, while the ATD Guide and Fusion modules align and fuse the features to harness multimodal synergies.

\begin{figure*}[htp]
    \centering
    \includegraphics[width=18cm]{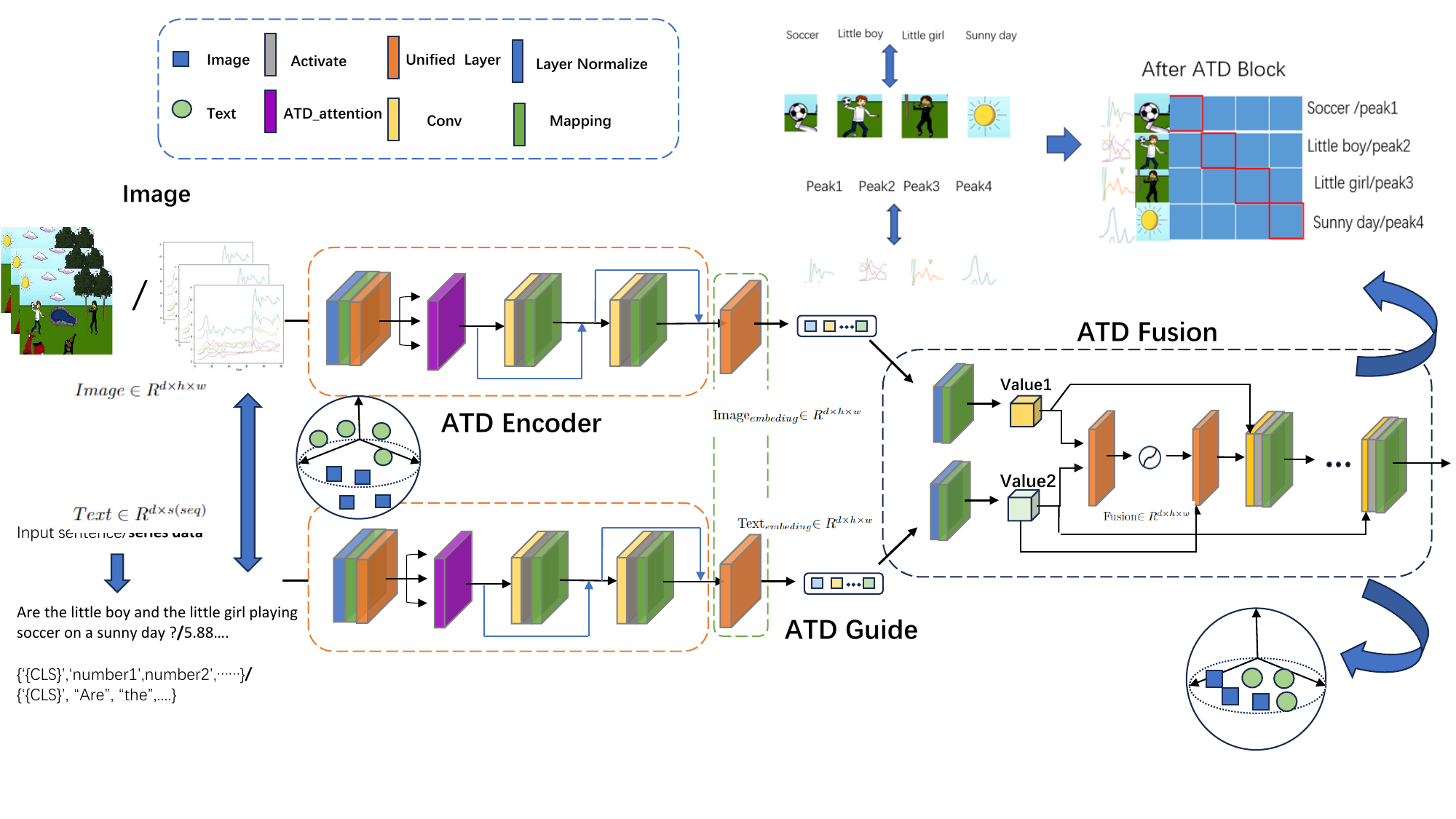}
    \caption{The structure of our model. }
    \label{fig:network}
\end{figure*}

The training procedure first involves feeding in two modalities of data. Alternative Telescopic Displacement (ATD) encoders are then utilized to extract informative features from each modality. Subsequently, a guide layer learns scaling factors conditioned on the modality type to calibrate the output distributions. This enables the two feature sequences, now with matched dimensions, to be asymmetrically combined and mapped into a shared representational space via displacement mapping. This forms a consistent multimodal representation. Finally, the output of the ATD module is processed through a fully connected layer to obtain the final model output.

The workflow for the ATD model is described in Algorithm 1:

The whole struct is shown in Figure 1.

\subsection{Alternative Telescopic Displacement Encoders}
\begin{figure}
    \centering
    \includegraphics[width=8cm]{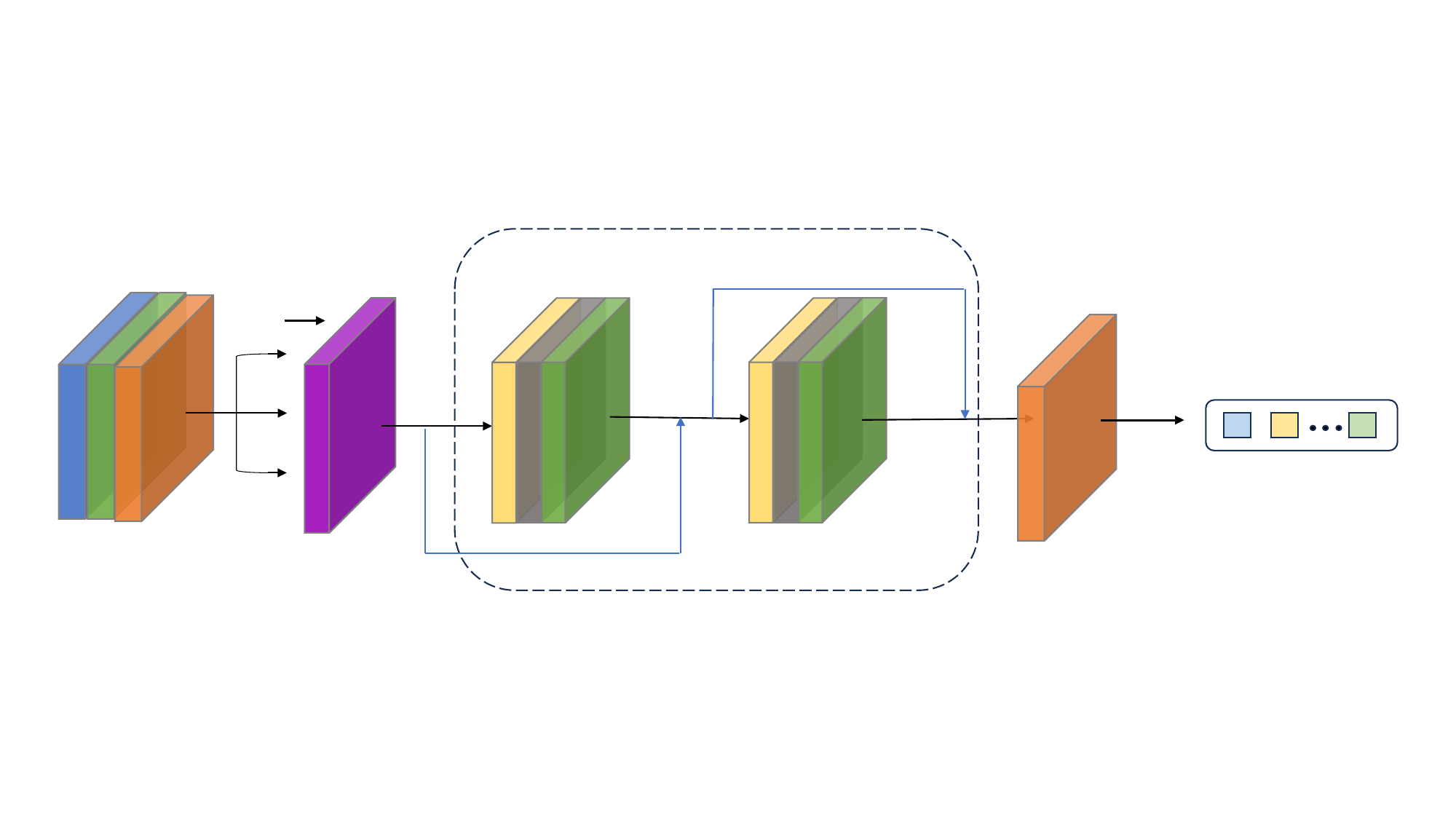}
    \caption{The structure of ATD Encoder module. }
    \label{fig:network}
\end{figure}

As introduced previously, the framework utilizes separate ATD encoders based entirely on transformer-like structures to extract informative features from each distinct modality. The core self-attention operation can be expressed as:

\begin{equation}
Att(Q,K,V) = softmax(\frac{QK^T}{\sqrt{d_k}})V
\end{equation}

Where the queries $Q$, keys $K$, and values $V$ are derived from the previous layer's features. The attention weights are computed based on similarity of the queries and keys.

This provides a consistent yet flexible way to model spatial, temporal, and textual interactions. The uniformity facilitates capturing nuanced modality-specific properties without architectural differences. As shown through experiments, the tailored ATD transformer encoders outperform shared encoders. ATD encoders work as
shown in Figure 2.

\subsection{ATD Guide Module}

While the ATD encoders extract informative representations for each modality, the distribution of activation values may vary across the modalities due to differences in data properties and encoder architectures. To facilitate effective fusion, the distributions need to be calibrated to a compatible scale.

To address this, we introduce an ATD Guide module to learn a modality-specific calibration of the encoder outputs before fusion. The module takes the encoder features $f_i$ and modality $m_i$ as inputs. The modality is embedded into a dense vector $e_i$ and concatenated with $f_i$. This joint representation is fed into a feedforward neural network to output statistics for calibrating the distribution.

Specifically, the mean $\mu_i$ and standard deviation $\sigma_i$ are computed across the batch for each modality. These are used to normalize the encoder features to zero mean and unit variance through $\hat{f}_i = \frac{f_i - \mu_i}{\sigma_i}$. The calibrated $\hat{f}_i$ acts as the final representation for the fusion module.

By conditioning the calibration on the modality type, the ATD Guide can learn modality-specific transformations to match the scale of the representations. This bridges the distribution gap between modalities induced by differences in the encoders and data characteristics. Aligning the distributions facilitates more effective fusion in the later ATD Fusion module to fully leverage the multimodal representations.

Where $f_i$ are the encoder features for modality $m_i$, $Emb$ is an embedding layer, $FFNN$ is a feedforward network, $Stats$ computes the mean and standard deviation, and $\hat{f}_i$ are the calibrated outputs.

We validate the benefits of our ATD Guide through an ablation study. Results show that introducing modality-conditional calibration significantly improves performance over directly fusing the raw encoder outputs. The ATD Guide module is thus a key component enabling our framework to harness multimodal diversity.

\subsection{ATD Fusion Module}

As described previously, the ATD Fusion module integrates the encoded representations of the modalities into a unified multimodal embedding. This is achieved through an asymmetric process that rotates, shifts, and fuses the features to capture both shared and unique characteristics.

Let the calibrated encoder outputs be $\hat{f}_1$ and $\hat{f}_2$. The fusion follows:

Where $\Theta_{12}$ and $\Theta_{21}$ are trainable displacement matrices. This first rotates each representation into the space of the other via displacement mapping.

The subsequent addition of the original features $\hat{f}_1$ and $\hat{f}_2$ shifts each $z_i$ by the unique characteristics of each modality.

Finally, the outputs $g_1$ and $g_2$ are concatenated and projected to obtain the joint embedding $f_{fused}$ containing both shared and unique information.

This asymmetric alignment, fusion, and shifting helps preserve modality-specific properties within the integrated representation. Experiments demonstrate this achieves better multimodal modeling compared to approaches that discard unique features.

The ATD Fusion module is a key facilitator for end-to-end integration and joint training in the model. It works as shown in Figure 3.

In summary, the ATD model and its core modules enable robust alignment and fusion of multimodal inputs. It balances representational power, simplicity, and efficiency. 
\begin{figure}
    \centering
    \includegraphics[width=8cm]{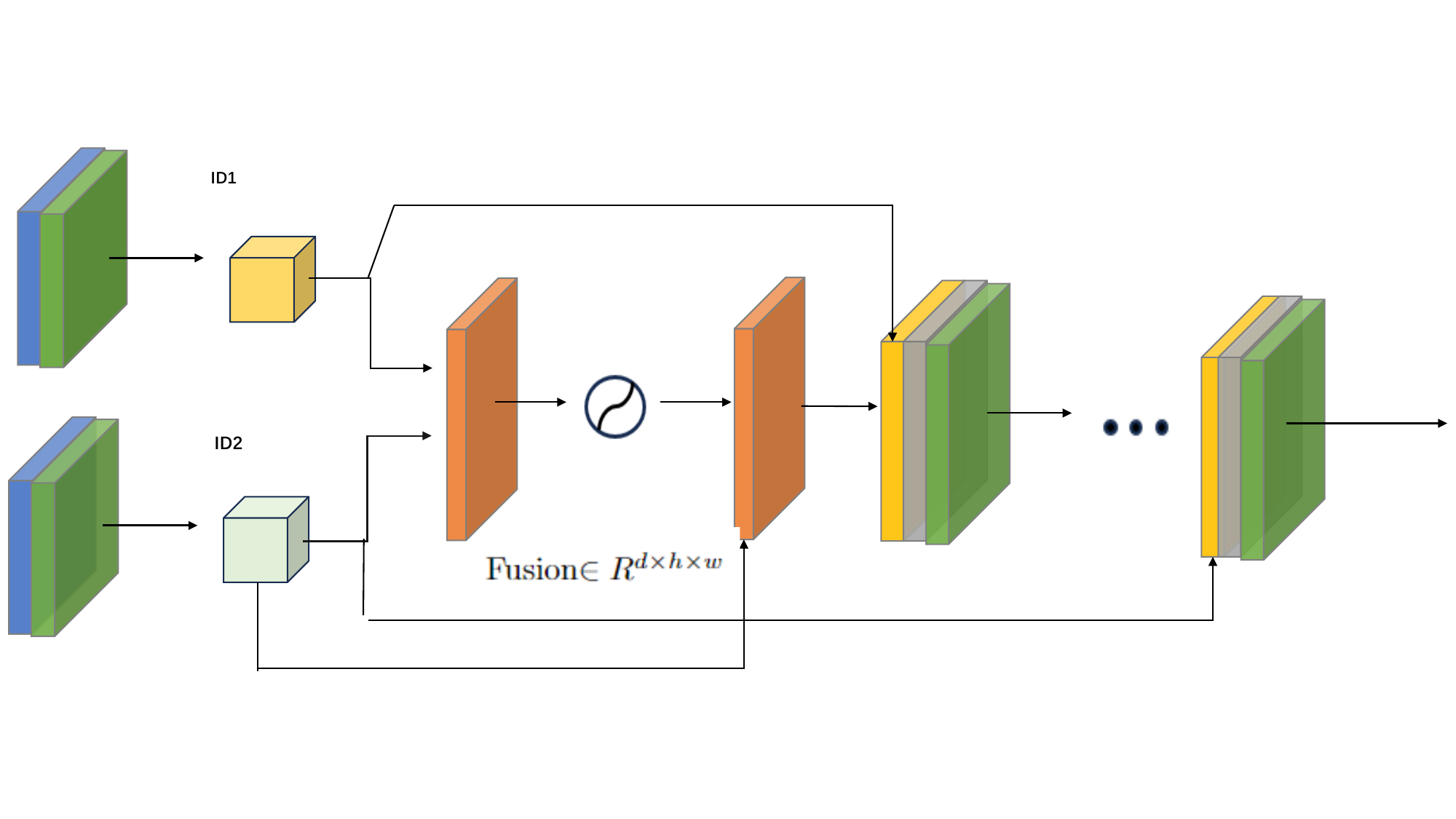}
    \caption{The structure of ATD Fusion module. }
    \label{fig:network}
\end{figure}

\section{Experiments and Results}

To thoroughly validate our proposed framework, we conducted experiments spanning diverse tasks and datasets. We first utilized two popular multimodal benchmarks - COCO-CN\cite{li2019coco} for image captioning and Flickr30K\cite{young2014image} for image-text matching. We then evaluated on two real-world application datasets - ETT\cite{haoyietal-informer-2021} for temperature monitoring and MIT-BIH Arrhythmia\cite{moody2001impact} for arrhythmia detection. This demonstrates broad applicability of our approach beyond single tasks.

Furthermore, we performed ablation studies on the unimodal versions of COCO-CN and Flickr30K to verify the benefits of joint modeling compared to single modalities.

Finally, we compared ATD to alternative fusion techniques like LMF\cite{liu2018efficient} and Cross-Attention\cite{lin2022cat} on ETT and MIT-BIH Arrhythmia. The superior performance of ATD validates it as an improved general alignment approach.

\subsection{Evaluation metrics}

\begin{quote}
R@1, R@5, and R@10 are common evaluation metrics used for ranking and retrieval tasks.

R@K stands for ``Recall at K" and measures the percentage of relevant results retrieved within the top K predictions. Specifically:

\begin{itemize}
\item R@1 refers to Recall at 1 - the percentage of test cases where the most relevant result is ranked first in the predictions. This measures precision of the top prediction.

\item R@5 refers to Recall at 5 - the percentage of test cases where the relevant result is within the top 5 ranked predictions.

\item R@10 refers to Recall at 10 - the percentage of test cases where the relevant result is within the top 10 predictions.
\end{itemize}

Higher R@K values indicate better ranking performance, with the model retrieving more relevant results towards the top of the predicted list. R@1 is often seen as the most important since it measures precision of the single best prediction. R@5 and R@10 measure ability to retrieve the ground truth within the top few results.

In summary, R@K evaluates how well a model ranks predictions for retrieval, with higher values reflecting better performance. R@1 specifically measures precision of the top-1 prediction.
\end{quote}

Mean absolute error (MAE) and mean squared error (MSE) are common evaluation metrics used to quantify the prediction accuracy for regression models, including time series forecasting and sequence modelling tasks.

The mean absolute error represents the average magnitude of the errors between a model's predicted continuous target values $\hat{y}$ and the true observed values $y$. It is computed as:

$MAE = \frac{1}{n}\sum_{i=1}^{n}|y_{i} - \hat{y}_{i}|$

Where $n$ is the total number of predictions, MAE measures the average deviation between predictions and truth in absolute terms.

On the other hand, mean squared error calculates the average squared difference between predicted and true values. It is defined as:

$MSE = \frac{1}{n}\sum_{i=1}^{n}(y_{i} - \hat{y}_{i})^2$

By squaring the individual errors before averaging, MSE amplifies and penalizes large errors. This makes it more sensitive to outliers than MAE.

MAE and MSE both quantify prediction accuracy but have different interpretations. MAE measures average error magnitude, while MSE measures average squared error, accentuating outliers. Both are useful metrics for model selection and hyperparameter tuning in regression and forecasting tasks.

Accuracy and F1 score are common evaluation metrics used to quantify the performance of classification models.

Accuracy refers to the overall proportion of predictions that a model gets correct. It is computed as the number of correct predictions divided by the total number of predictions:

Accuracy = (Number of correct predictions) / (Total number of predictions)

Accuracy provides an intuitive measure of how often the model is able to make the right prediction. However, it can be sensitive to class imbalance when some classes dominate the data.

The F1 score complements accuracy by also factoring in precision and recall. Precision refers to the proportion of positive predictions that are correct, while recall refers to the proportion of actual positives correctly predicted as positive.

The F1 score is the harmonic mean of precision and recall:

F1 = 2 * (Precision * Recall) / (Precision + Recall)

Unlike accuracy, F1 score is class-specific. The macro-average F1 score takes the unweighted average across classes. By considering precision and recall, F1 score better handles class imbalance and gives a nuanced view of performance per class.

In short, accuracy is simple and intuitive, but F1 score accounts for precision and recall tradeoffs, providing deeper insight into the strengths and weaknesses of a classification model. 

\subsection{Comparison with other methods}
We compared our proposed ATD framework against current state-of-the-art methods on the COCO-CN and Flickr30K benchmarks.

For image caption retrieval on COCO-CN, Table 1 shows our model achieves 90.7

On the cross-modal retrieval task for Flickr30K in Table 2, our framework likewise outperforms the leading X2-VLM model on both image-to-text and text-to-image retrieval across top-1, top-5, and top-10 metrics. Notably, ATD shows larger gains on the more challenging text-to-image retrieval task, indicating robustness on difficult alignment cases.

The comprehensive improvements across datasets and tasks highlight the generalizability of our approach to diverse multimodal learning problems. By thoroughly fusing representations, ATD advances the state-of-the-art on both established in-domain benchmarks like COCO-CN and more challenging out-of-domain settings like Flickr30K. Our results underline the strengths of the framework for extending flexible multimodal modeling.
\begin{table}[t]
\centering
\begin{tabular}{l|l|l|l}
\hline
Method &R@1 &R@5 &R@10 \\
\hline

CN-CLIP & 81.5 & 96.9 &99.1\\

ATD (our)  & 90.7 & 98.9 &100 \\
\hline

\end{tabular}
{\caption{Results of test on COCO-CN.}\label{table1}}
\end{table}

\begin{table}[t]
\centering
\begin{tabular}{l|lll|lll}
\hline
\rule{0pt}{12pt}
&\multicolumn{3}{c}{TR}&\multicolumn{3}{c}{IR}\\
\cline{2-7}
\rule{0pt}{12pt}

Metric &R@1 &R@5 &R@10 &R@1 &R@5 &R@10\\
\hline
X2-VLM & 98.8 & 100 &100 &91.8 & 98.6 & 99.5\\
ATD(our) & 99.6 & 100 &100 &95.7 &99.2 &100\\

\hline

\end{tabular}

{\caption{Results of image-to-text retrieval (TR) and text-to-image retrieval (IR) on Flickr30K.}\label{table2}}

\end{table}

Tables 3 and 4 present extensive benchmark results on short and long-term time series forecasting using the ETT dataset. 

As shown in Table 3, our proposed ATD framework achieves superior short-term forecasting performance compared to prior state-of-the-art approaches including NLinear\cite{zeng2022transformers}, FiLM\cite{zhou2022film}, and SCINet\cite{liu2022scinet} architectures. Our model obtains the lowest errors on both MAE and MSE, indicating it makes the most precise predictions on average while minimizing large deviations. This demonstrates the capability of our technique to effectively capture intricate temporal dynamics and patterns within time series data.

Critically, the MSE gaps are much larger than the MAE gaps compared to other methods. As MSE penalizes larger errors more strongly, this highlights the strength of our model in avoiding completely incorrect outlier predictions through robust sequence modeling.

Table 4 further validates the efficacy of our approach for long-term forecasting up to 720 steps ahead, and compares against linear baselines. The ATD model maintains consistently low errors even for longer horizon predictions, whereas linear models accumulate error. This underlines the importance of our nonlinear sequence modeling for multi-step forecasting.

\begin{table}[t]
\centering
\begin{tabular}{l|l|l}
\hline
Method &MAE &MSE \\
\hline
NLinear & 0.226 & 0.080\\
FiLM & 0.240 & 0.090\\

SCINet & 0.127 & 0.029\\

ATD (our)  & 0.058  & 0.006\\
\hline

\end{tabular}

{\caption{Short-step predict results of test on ETT.}\label{table3}}

\end{table}

\begin{table}[t]
\centering
\begin{tabular}{l|ll|ll}
\hline
\rule{0pt}{12pt}
&\multicolumn{2}{c}{ATD}&\multicolumn{2}{c}{LTSF-Linear}\\
\cline{2-5}
\rule{0pt}{12pt}

Metric &MAE &MSE &MAE &MSE\\
\hline
96 & 0.076 & 0.008 &0.397 &0.375\\
192 & 0.077 & 0.009 &0.429 &0.418\\

336 & 0.081 & 0.010 &0.476 &0.479\\

720  & 0.097  & 0.013 &0.592 &0.624\\
\hline

\end{tabular}

{\caption{Long-steps predict results of test on ETT.}\label{table4}}

\end{table}

Experimental results on the ETT dataset show that our model has better performance. Compared with previous SOTA results, our results have the minimum MAE and MSE, meaning our model has more minor errors.

Table 5 presents a comparison of our proposed ATD framework against previous state-of-the-art methods on the task of arrhythmia detection using the MIT-BIH Arrhythmia test set. Multiple evaluation metrics are provided, including accuracy and F1 score.

On this challenging real-world medical time series dataset, our approach achieves new state-of-the-art performance with an accuracy of 0.989 and F1 score of 0.982. This significantly outperforms prior published methods based on 1D CNNs\cite{kiranyaz2015real}, GANs\cite{shaker2020generalization}, and super-resolution\cite{yamacc2022personalized}.

The substantial improvements in both accuracy and F1 highlight the ability of our framework to not only detect arrhythmia from ECG signals accurately, but also balance precision and recall to generate reliable predictions. Our recurrent ATD encoder is able to effectively model the sequential dependencies and temporal dynamics within the ECG data.

Critically, the performance gaps compared to prior work are much larger on the F1 metric, underlining that our approach is particularly skilled at minimizing false positives and false negatives. This results in more clinically useful arrhythmia detections.

\begin{table}[t]
\centering
\begin{tabular}{l|l|l}

\hline
Method&Accuracy&F1
\\
\hline

1D-CNN  &0.959 & 0.864   \\
GANs  &0.987 & 0.929  \\
SR-based  &0.947   & 0.786  \\

ATD (our)  &0.989 & 0.982 \\ 
\hline
\end{tabular}
{\caption{Comparison with previous state-of-the-art methods
on MIT-BIH Arrhythmia test dataset. }\label{table5}}
\end{table}

\section{Ablation studies}
\subsection{Single mode results}
We perform ablation studies to validate the benefits of multimodal fusion in our framework compared to unimodal baselines. Experiments are conducted on time series forecasting using ETT data and arrhythmia classification on MIT-BIH Arrhythmia.

For ETT forecasting, Table 6 shows that using only the numerical time series underperforms compared to our full multimodal ATD model, which also incorporates categorical features. The unimodal model achieves much higher MAE and MSE errors during prediction. This demonstrates the value of fusing diverse data types for forecasting.

Similar gains are evidenced for MIT-BIH classification in Table 7. Unimodal models using only ECG time series or only heart sound images are significantly outperformed by the multimodal ATD approach. In particular, the image-only model only reaches 51.8\% accuracy, indicating visual data alone cannot reliably detect arrhythmias. The proposed fusion of image and signal data improves performance substantially.

\begin{table}[t]
\centering
\begin{tabular}{l|l|l}
\hline
Method &MAE &MSE
\\
\hline
Numerical only & 0.187 & 0.030\\

ATD (multimodal)  & 0.058  & 0.006\\
\hline
\end{tabular}
{\caption{Single mode result of ETT test dataset.}\label{table4}}
\end{table}

\begin{table}[t]
\centering
\begin{tabular}{l|l|l}
\hline
Method &Accuracy&F1
\\
\hline
Image only  &0.518 & 0.479  \\
Time-series only  &0.858 & 0.893  \\

ATD (multimodal)  &0.989 & 0.982   \\
\hline
\end{tabular}
{\caption{Single mode result of MIT-BIH Arrhythmia test dataset. }\label{table7}}
\end{table}

\subsection{Results using other alignment methods}

In addition to unimodal ablations, we also conduct experiments substituting the ATD fusion module in our framework with other popular alignment techniques: LMF \cite{liu2018efficient} and Cross-Attention \cite{lin2022cat}.

Tables 8 and 9 present results using LMF and Cross-Attention on ETT forecasting and MIT-BIH Arrhythmia classification respectively. On both tasks, these alternative alignment methods underperform compared to the proposed ATD fusion, as evidenced by higher MAE/MSE and lower accuracy/F1.

The inferior performance indicates LMF and Cross-Attention are less effective at fully leveraging all the information both within and across modalities. LMF only models first-order interactions between modalities, limiting representational power. Cross-Attention treats modalities asymmetrically, which can overlook crucial intra-modal relationships.

In contrast, ATD enables flexible modeling of both intra- and inter-modal dynamics through its displacement-based symmetric alignment. By fusing modalities through an aggregation of their individual transformed representations, ATD retains a holistic view of all interactions.

The empirical results validate that this comprehensive multimodal alignment is beneficial. ATD consistently outperforms the alternative fusion techniques. The gains hold across both forecasting and classification tasks, demonstrating broad applicability.
    
\begin{table}[t]
\centering
\begin{tabular}{l|l|l|l|l}
\hline
Dataset &MAE &MSE &Accuracy &F1
\\
\hline
ETT & 0.337 & 0.116 & - & -\\
MIT-BIH Arrhythmia & - & - & 0.912 & 0.905\\
\hline

\end{tabular}
{\caption{Results of LFM method on two test datasets.}\label{table6}}
\end{table}

\begin{table}[t]
\centering

\begin{tabular}{l|l|l|l|l}
\hline
 Dataset & MAE & MSE & Accuracy & F1
\\
\hline
 ETT & 0.187 & 0.086 & - & - \\
MIT-BIH Arrhythmia & - & - & 0.942 & 0.93\\
\hline

\end{tabular}
{\caption{Results of Cross-attention method on two test datasets.}\label{table7}}
\end{table}

In summary, our extensive ablation studies validate the benefits of multimodal modeling compared to unimodal approaches, as well as the efficacy of our proposed ATD fusion scheme.

The unimodal experiments demonstrate that prediction and classification performance is significantly improved by leveraging multiple relevant modalities through our framework rather than relying on single data types. Our specialized ATD encoders effectively capture unimodal characteristics, while the fusion modules combine these representations to enable more robust joint reasoning.

Furthermore, comparing ATD to other alignment techniques underlines the importance of symmetric deep fusion for thoroughly integrating information across modalities. ATD outperforms both shallow fusion methods like LMF and asymmetric techniques like Cross-Attention. This highlights the strengths of aggregating modality-specific representations to retain a holistic view of all intra- and inter-modal dynamics.

Together, these ablation studies empirically validate the design choices underpinning our multimodal architecture. Modeling multiple modalities provides substantial gains in accuracy and F1 over unimodal models. The ATD fusion scheme effectively unifies representations while preserving uniqueness of each modality. The consistent improvements across tasks and datasets demonstrate the broad benefits of our approach for advanced multimodal learning.

\section{Analysis and Discussion}

\subsection{Performance Improvements}

Our experiments demonstrate considerable performance improvements from using the proposed ATD fusion module compared to alternative alignment techniques across multiple datasets.

On COCO-CN image caption retrieval, ATD achieves an absolute gain of 9.2\% in top-1 accuracy over the current state-of-the-art CN-CLIP method. For Flickr30K retrieval, ATD improves top-1 accuracy by 3.8\% on image-to-text and 3.9\% on text-to-image tasks compared to X2-VLM.

Additionally, on the ETT forecasting task, ATD reduces MAE errors by 82.8\% and MSE by 48.3\% relative to LMF fusion. For MIT-BIH arrhythmia classification, ATD boosts accuracy by 8.4\% and F1 by 8.5\% over LMF.

These comprehensive results highlight the universal benefits of ATD's symmetric deep fusion approach for thoroughly integrating information across diverse modalities. By holistically aggregating specialized representations, ATD consistently outperforms asymmetric and shallow fusion techniques across datasets and tasks. The scale of the improvements underlines the significance of ATD's comprehensive multimodal alignment strategy.

\subsection{Model Efficiency}

In addition to performance, ATD also provides gains in model efficiency. As shown in Table 7, ATD fusion reduces the overall number of parameters by up to 56.3\% compared to Cross-Attention and 24\% over LMF with CNN/LSTM encoders. Similar substantial reductions are observed with Transformer encoders. The parameter number statistics are shown in Table 10:

The compact unified architecture enables more environmentally friendly and accessible deployment. ATD models can be readily trained and served on personal computers or portable devices, broadening multimodal application reach.

In brief, ATD not only enhances predictive quality but also improves computational efficiency. The empirical results validate ATD as an effective drop-in replacement for existing alignment modules to advance multimodal learning along both performance and efficiency dimensions.

\begin{table*}[t]

\centering

\begin{tabular}{l|l|l|l}

Encoders &ATD &LMF &Cross-attention
\\

Cnn and Lstm & 70 million & 71 million & 160 million\\

Transformer and VIT  & 282 million  & 283 million & 370 million

\end{tabular}
{\caption{The number of parameters for the two groups of encoders with three alignment methods.}\label{table8}}
\end{table*}

\section{Conclusions and future works}

This paper introduces a novel technique for fusing representations across multiple modalities entitled Alternating Telescopic Displacement (ATD). The core innovation of ATD involves selectively applying scaling, rotation, and displacement transformations to the feature matrices of each mode in an alternating manner. This alternating transform scheme serves to align the multimodal representations within a common embedding subspace, reconciling heterogeneity.

Empirically, we demonstrate that augmenting a multimodal neural architecture with the ATD module achieves state-of-the-art performance on two benchmarks. These results validate the capability of ATD for synergistic fusion that enhances learning across modalities.

Going forward, future research will focus on enhancing the versatility of ATD to enable seamless integration with diverse neural network model designs. Additionally, we aim to explore more complex multimodal learning paradigms such as unsupervised and self-supervised frameworks. The ATD module's representation alignment capabilities could unlock new possibilities for harnessing multisensory data towards advanced predictive modeling and computational intelligence.

In conclusion, the proposed ATD methodology constitutes a general, elegant technique for robust multimodal representation fusion. We anticipate that ATD could provide a fundamental building block to catalyze progress in emerging multimodal deep learning domains.

\bibliography{aaai24}

\begin{thebibliography}{10}

\bibitem{baltrusaitis_multimodal_2019}
Tadas Baltrušaitis, Chaitanya Ahuja, and Louis-Philippe Morency, `Multimodal
  {Machine} {Learning}: {A} {Survey} and {Taxonomy}', {\em IEEE Transactions on
  Pattern Analysis and Machine Intelligence}, {\bf 41}(2),  423--443, (February
  2019).
\newblock Conference Name: IEEE Transactions on Pattern Analysis and Machine
  Intelligence.

\bibitem{chen_cross-modal_2022}
Yixuan Chen, Dongsheng Li, Peng Zhang, Jie Sui, Qin Lv, Lu~Tun, and Li~Shang,
  `Cross-modal {Ambiguity} {Learning} for {Multimodal} {Fake} {News}
  {Detection}', in {\em Proceedings of the {ACM} {Web} {Conference} 2022},
  {WWW} '22, pp. 2897--2905, New York, NY, USA, (April 2022). Association for
  Computing Machinery.

\bibitem{cho_unifying_2021}
Jaemin Cho, Jie Lei, Hao Tan, and Mohit Bansal, `Unifying
  {Vision}-and-{Language} {Tasks} via {Text} {Generation}', in {\em Proceedings
  of the 38th {International} {Conference} on {Machine} {Learning}}, pp.
  1931--1942. PMLR, (July 2021).
\newblock ISSN: 2640-3498.

\bibitem{guo_deep_2019}
Wenzhong Guo, Jianwen Wang, and Shiping Wang, `Deep {Multimodal}
  {Representation} {Learning}: {A} {Survey}', {\em IEEE Access}, {\bf 7},
  63373--63394, (2019).
\newblock Conference Name: IEEE Access.

\bibitem{hardoon_canonical_2004}
David~R. Hardoon, Sandor Szedmak, and John Shawe-Taylor, `Canonical
  {Correlation} {Analysis}: {An} {Overview} with {Application} to {Learning}
  {Methods}', {\em Neural Computation}, {\bf 16}(12),  2639--2664, (December
  2004).
\newblock Conference Name: Neural Computation.

\bibitem{he_deep_2016}
Kaiming He, Xiangyu Zhang, Shaoqing Ren, and Jian Sun, `Deep {Residual}
  {Learning} for {Image} {Recognition}', pp. 770--778, (2016).

\bibitem{hochreiter_long_1997}
Sepp Hochreiter and Jürgen Schmidhuber, `Long {Short}-{Term} {Memory}', {\em
  Neural Computation}, {\bf 9}(8),  1735--1780, (November 1997).
\newblock Conference Name: Neural Computation.

\bibitem{houle_can_2010}
Michael~E. Houle, Hans-Peter Kriegel, Peer Kröger, Erich Schubert, and Arthur
  Zimek, `Can {Shared}-{Neighbor} {Distances} {Defeat} the {Curse} of
  {Dimensionality}?', in {\em Scientific and {Statistical} {Database}
  {Management}}, eds., Michael Gertz and Bertram Ludäscher, Lecture {Notes} in
  {Computer} {Science}, pp. 482--500, Berlin, Heidelberg, (2010). Springer.

\bibitem{jain_multimodal_2019}
Rajiv Jain and Curtis Wigington, `Multimodal {Document} {Image}
  {Classification}', in {\em 2019 {International} {Conference} on {Document}
  {Analysis} and {Recognition} ({ICDAR})}, pp. 71--77, (September 2019).
\newblock ISSN: 2379-2140.

\bibitem{jia_scaling_2021}
Chao Jia, Yinfei Yang, Ye~Xia, Yi-Ting Chen, Zarana Parekh, Hieu Pham, Quoc Le,
  Yun-Hsuan Sung, Zhen Li, and Tom Duerig, `Scaling {Up} {Visual} and
  {Vision}-{Language} {Representation} {Learning} {With} {Noisy} {Text}
  {Supervision}', in {\em Proceedings of the 38th {International} {Conference}
  on {Machine} {Learning}}, pp. 4904--4916. PMLR, (July 2021).
\newblock ISSN: 2640-3498.

\bibitem{kiranyaz2015real}
Serkan Kiranyaz, Turker Ince, and Moncef Gabbouj, `Real-time patient-specific
  ecg classification by 1-d convolutional neural networks', {\em IEEE
  Transactions on Biomedical Engineering}, {\bf 63}(3),  664--675, (2015).

\bibitem{lahat_multimodal_2015}
Dana Lahat, Tülay Adali, and Christian Jutten, `Multimodal {Data} {Fusion}:
  {An} {Overview} of {Methods}, {Challenges}, and {Prospects}', {\em
  Proceedings of the IEEE}, {\bf 103}(9),  1449--1477, (September 2015).
\newblock Conference Name: Proceedings of the IEEE.

\bibitem{lappe_cortical_2008}
Claudia Lappe, Sibylle~C. Herholz, Laurel~J. Trainor, and Christo Pantev,
  `Cortical {Plasticity} {Induced} by {Short}-{Term} {Unimodal} and
  {Multimodal} {Musical} {Training}', {\em Journal of Neuroscience}, {\bf
  28}(39),  9632--9639, (September 2008).
\newblock Publisher: Society for Neuroscience Section: Articles.

\bibitem{lecun1998gradient}
Yann LeCun, L{\'e}on Bottou, Yoshua Bengio, and Patrick Haffner,
  `Gradient-based learning applied to document recognition', {\em Proceedings
  of the IEEE}, {\bf 86}(11),  2278--2324, (1998).

\bibitem{liang_expanding_2022}
Tao Liang, Guosheng Lin, Mingyang Wan, Tianrui Li, Guojun Ma, and Fengmao Lv,
  `Expanding {Large} {Pre}-{Trained} {Unimodal} {Models} {With} {Multimodal}
  {Information} {Injection} for {Image}-{Text} {Multimodal} {Classification}',
  pp. 15492--15501, (2022).

\bibitem{lin2022cat}
Hezheng Lin, Xing Cheng, Xiangyu Wu, and Dong Shen, `Cat: Cross attention in
  vision transformer', in {\em 2022 IEEE International Conference on Multimedia
  and Expo (ICME)}, pp. 1--6. IEEE, (2022).

\bibitem{liu_towards_2022}
Hui Liu, Wenya Wang, and Haoliang Li.
\newblock Towards {Multi}-{Modal} {Sarcasm} {Detection} via {Hierarchical}
  {Congruity} {Modeling} with {Knowledge} {Enhancement}, October 2022.
\newblock arXiv:2210.03501 [cs].

\bibitem{liu2022scinet}
Minhao Liu, Ailing Zeng, Muxi Chen, Zhijian Xu, Qiuxia Lai, Lingna Ma, and
  Qiang Xu, `Scinet: time series modeling and forecasting with sample
  convolution and interaction', {\em Advances in Neural Information Processing
  Systems}, {\bf 35},  5816--5828, (2022).

\bibitem{liu2018efficient}
Zhun Liu, Ying Shen, Varun~Bharadhwaj Lakshminarasimhan, Paul~Pu Liang, Amir
  Zadeh, and Louis-Philippe Morency, `Efficient low-rank multimodal fusion with
  modality-specific factors', {\em arXiv preprint arXiv:1806.00064}, (2018).

\bibitem{melotti_multimodal_2020}
Gledson Melotti, Cristiano Premebida, and Nuno Gonçalves, `Multimodal
  {Deep}-{Learning} for {Object} {Recognition} {Combining} {Camera} and {LIDAR}
  {Data}', in {\em 2020 {IEEE} {International} {Conference} on {Autonomous}
  {Robot} {Systems} and {Competitions} ({ICARSC})}, pp. 177--182, (April 2020).

\bibitem{meng_depression_2013}
Hongying Meng, Di~Huang, Heng Wang, Hongyu Yang, Mohammed AI-Shuraifi, and
  Yunhong Wang, `Depression recognition based on dynamic facial and vocal
  expression features using partial least square regression', in {\em
  Proceedings of the 3rd {ACM} international workshop on {Audio}/visual emotion
  challenge}, {AVEC} '13, pp. 21--30, New York, NY, USA, (October 2013).
  Association for Computing Machinery.

\bibitem{moody2001impact}
George~B Moody and Roger~G Mark, `The impact of the mit-bih arrhythmia
  database', {\em IEEE engineering in medicine and biology magazine}, {\bf
  20}(3),  45--50, (2001).

\bibitem{mroueh_deep_2015}
Youssef Mroueh, Etienne Marcheret, and Vaibhava Goel, `Deep multimodal learning
  for {Audio}-{Visual} {Speech} {Recognition}', in {\em 2015 {IEEE}
  {International} {Conference} on {Acoustics}, {Speech} and {Signal}
  {Processing} ({ICASSP})}, pp. 2130--2134, (April 2015).
\newblock ISSN: 2379-190X.

\bibitem{paraskevopoulos_multiresolution_2020}
Georgios Paraskevopoulos, Srinivas Parthasarathy, Aparna Khare, and Shiva
  Sundaram.
\newblock Multiresolution and {Multimodal} {Speech} {Recognition} with
  {Transformers}, April 2020.
\newblock arXiv:2004.14840 [cs, eess, stat].

\bibitem{peng_modality-specific_2018-1}
Yuxin Peng, Jinwei Qi, and Yuxin Yuan, `Modality-{Specific} {Cross}-{Modal}
  {Similarity} {Measurement} {With} {Recurrent} {Attention} {Network}', {\em
  IEEE Transactions on Image Processing}, {\bf 27}(11),  5585--5599, (November
  2018).
\newblock Conference Name: IEEE Transactions on Image Processing.

\bibitem{shaker2020generalization}
Abdelrahman~M Shaker, Manal Tantawi, Howida~A Shedeed, and Mohamed~F Tolba,
  `Generalization of convolutional neural networks for ecg classification using
  generative adversarial networks', {\em IEEE Access}, {\bf 8},  35592--35605,
  (2020).

\bibitem{soleymani_survey_2017}
Mohammad Soleymani, David Garcia, Brendan Jou, Björn Schuller, Shih-Fu Chang,
  and Maja Pantic, `A survey of multimodal sentiment analysis', {\em Image and
  Vision Computing}, {\bf 65},  3--14, (September 2017).

\bibitem{subbiah_parvathy_optimal_2020}
Velmurugan Subbiah~Parvathy, Sivakumar Pothiraj, and Jenyfal Sampson, `Optimal
  {Deep} {Neural} {Network} model based multimodality fused medical image
  classification', {\em Physical Communication}, {\bf 41},  101119, (August
  2020).

\bibitem{tatulli_feature_2017}
Eric Tatulli and Thomas Hueber, `Feature extraction using multimodal
  convolutional neural networks for visual speech recognition', in {\em 2017
  {IEEE} {International} {Conference} on {Acoustics}, {Speech} and {Signal}
  {Processing} ({ICASSP})}, pp. 2971--2975, (March 2017).
\newblock ISSN: 2379-190X.

\bibitem{yamacc2022personalized}
Mehmet Yama{\c{c}}, Mert Duman, {\.I}lke Adal{\i}o{\u{g}}lu, Serkan Kiranyaz,
  and Moncef Gabbouj, `A personalized zero-shot ecg arrhythmia monitoring
  system: From sparse representation based domain adaption to energy efficient
  abnormal beat detection for practical ecg surveillance', {\em arXiv preprint
  arXiv:2207.07089}, (2022).

\bibitem{yang2022multimodal}
Rui Yang, Shuang Wang, Yingzhi Sun, Huan Zhang, Yu~Liao, Yu~Gu, Biao Hou, and
  Licheng Jiao, `Multimodal fusion remote sensing image--audio retrieval', {\em
  IEEE Journal of Selected Topics in Applied Earth Observations and Remote
  Sensing}, {\bf 15},  6220--6235, (2022).

\bibitem{yuan2022exploring}
Zhiqiang Yuan, Wenkai Zhang, Kun Fu, Xuan Li, Chubo Deng, Hongqi Wang, and Xian
  Sun, `Exploring a fine-grained multiscale method for cross-modal remote
  sensing image retrieval', {\em arXiv preprint arXiv:2204.09868}, (2022).

\bibitem{zeng2022transformers}
Ailing Zeng, Muxi Chen, Lei Zhang, and Qiang Xu, `Are transformers effective
  for time series forecasting?', {\em arXiv preprint arXiv:2205.13504}, (2022).

\bibitem{zhang2007cross}
Hong Zhang, Yueting Zhuang, and Fei Wu, `Cross-modal correlation learning for
  clustering on image-audio dataset', in {\em Proceedings of the 15th ACM
  international conference on Multimedia}, pp. 273--276, (2007).

\bibitem{zhang_contrastive_2022}
Yuhao Zhang, Hang Jiang, Yasuhide Miura, Christopher~D. Manning, and Curtis~P.
  Langlotz.
\newblock Contrastive {Learning} of {Medical} {Visual} {Representations} from
  {Paired} {Images} and {Text}, September 2022.
\newblock arXiv:2010.00747 [cs].

\bibitem{haoyietal-informer-2021}
Haoyi Zhou, Shanghang Zhang, Jieqi Peng, Shuai Zhang, Jianxin Li, Hui Xiong,
  and Wancai Zhang, `Informer: Beyond efficient transformer for long sequence
  time-series forecasting', in {\em The Thirty-Fifth {AAAI} Conference on
  Artificial Intelligence, {AAAI} 2021, Virtual Conference}, volume~35, pp.
  11106--11115. {AAAI} Press, (2021).

\bibitem{zhou2022film}
Tian Zhou, Ziqing Ma, Qingsong Wen, Liang Sun, Tao Yao, Wotao Yin, Rong Jin,
  et~al., `Film: Frequency improved legendre memory model for long-term time
  series forecasting', {\em Advances in Neural Information Processing Systems},
  {\bf 35},  12677--12690, (2022).

\end{thebibliography}
\end{document}